\documentclass[conference]{IEEEtran}
\IEEEoverridecommandlockouts
\usepackage{cite}
\usepackage{float}
\usepackage{caption} 
\usepackage{balance}
\usepackage{amsmath,amssymb,amsfonts}
\usepackage{algorithmic}
\usepackage{graphicx}
\usepackage{textcomp}
\usepackage{xcolor}
\usepackage{todonotes}
\usepackage{tikz}

\def\BibTeX{{\rm B\kern-.05em{\sc i\kern-.025em b}\kern-.08em
    T\kern-.1667em\lower.7ex\hbox{E}\kern-.125emX}}
\IEEEoverridecommandlockouts
\begin{document}

\title{Transformer-Based Cognitive Radio: Adaptive Modulation Strategies Using Transformer Models\\

}

\author{
    \IEEEauthorblockN{Andrea Melis}
    \IEEEauthorblockA{
        Dpt. of Eng. and Computer Science\\
        University of Bologna\\
        Bologna, Italy\\
        a.melis@unibo.it}
    \and
    \IEEEauthorblockN{Andrea Piroddi}
    \IEEEauthorblockA{
        Dpt. of Eng. and Computer Science\\
        University of Bologna\\
        Bologna, Italy\\
        andrea.piroddi@unibo.it}
    \and
    \IEEEauthorblockN{Roberto Girau}
    \IEEEauthorblockA{
        Dpt. of Eng. and Computer Science\\
        University of Bologna\\
        Bologna, Italy\\
        roberto.girau@unibo.it}
}

\maketitle

\begin{tikzpicture}[remember picture, overlay]
  \node[anchor=north east, xshift=-1.5cm, yshift=-0.5cm] at (current page.north east) {
    \begin{minipage}{0.6\textwidth}
    \raggedleft
    \footnotesize
    V. International Conference on Electrical, Computer and Energy Technologies (ICECET 2025)\\
    3--6 July 2025, Paris, France
    \end{minipage}
  };
\end{tikzpicture}

\begin{tikzpicture}[remember picture, overlay]
  \node[anchor=south west, xshift=1.5cm, yshift=0.75cm] at (current page.south west) {
    \footnotesize 979-8-3315-3559-9/25/\$31.00~\copyright~2025 IEEE
  };
\end{tikzpicture}

\begin{abstract}
Cognitive Radio (CR) systems, which dynamically adapt to changing spectrum environments, could benefit significantly from advancements in machine learning technologies. These systems can be enhanced in terms of spectral efficiency, robustness, and security through innovative approaches such as the use of Transformer models. This work investigates the application of Transformer models, specifically the GPT-2 architecture, to generate novel modulation schemes for wireless communications. By training a GPT-2 model on a dataset of existing modulation formulas, new modulation schemes has been created. These generated schemes are then compared to traditional methods using key performance metrics such as Signal-to-Noise Ratio (SNR) and Power Spectrum Density (PSD). The results show that Transformer-generated modulation schemes can achieve performance comparable to, and in some cases outperforming, traditional methods. This demonstrates that advanced CR systems could greatly benefit from the implementation of Transformer models, leading to more efficient, robust, and secure communication systems.
\end{abstract}

\begin{IEEEkeywords}
Transformer Model, Cognitive Radio, Modulation Schemes, Wireless Communications, Signal Processing
\end{IEEEkeywords}
\balance
\section{Introduction}
Modulation schemes play a pivotal role in wireless communications, determining how information is encoded onto carrier signals for transmission over the airwaves. Efficient modulation techniques are essential for maximizing spectral efficiency, improving signal robustness, and minimizing interference in increasingly crowded spectral environments. Traditional modulation methods, such as Amplitude Modulation (AM), Frequency Modulation (FM), and Phase Modulation (PM), have been manually designed and optimized over decades. However, the increasing complexity and demand for higher data rates in modern communication systems pose significant challenges for these conventional approaches.

Designing efficient modulation schemes involves balancing various factors, including bandwidth efficiency, power efficiency, and resilience to noise and interference. This process typically requires extensive domain expertise and iterative testing, which can be time-consuming and resource-intensive. Furthermore, traditional modulation techniques may not fully exploit the potential of modern computational methods and machine learning algorithms to discover innovative and optimized solutions.

Recent advancements in machine learning, particularly with Transformer models such as the Generative Pre-trained Transformer (GPT), offer promising new avenues for automatic generation and optimization of modulation schemes. Transformers, known for their ability to capture complex patterns and generate coherent sequences, have demonstrated remarkable success in natural language processing and other sequential data tasks. Applying these models to the domain of signal processing and modulation design could revolutionize the way new schemes are conceived and implemented.

Cognitive Radio (CR) systems, which dynamically adapt to changing spectrum environments to improve the efficiency of spectrum usage, stand to benefit significantly from such advancements. The integration of machine learning techniques in CR systems has already enhanced capabilities in spectrum sensing, modulation classification, and adaptive modulation. Recent studies have shown that deep learning models, including Transformers, can significantly improve the robustness and accuracy of spectrum sensing, a critical component of CR systems, by detecting unused spectrum bands more effectively\cite{khalek2023advances}. Furthermore, the innovative use of Transformer models in cognitive radio systems can enhance security applications. Adversarial modulation schemes, for example, can create secure communication channels that are difficult for adversaries to predict, intercept, or decode, thereby enhancing data confidentiality and integrity \cite{wang2023adversarial}. These techniques can also be integrated into intrusion detection systems to identify and counteract unauthorized modulation patterns, strengthening overall network defenses \cite{kim2021deep}.




This paper investigates the use of Transformer models, specifically the GPT-2 architecture, to generate novel modulation schemes for wireless communications. We train the GPT-2 model on a dataset of existing modulation formulas, enabling it to generate new schemes. These generated schemes are assessed under identical conditions using key performance metrics such as Signal-to-Noise Ratio (SNR) and Power Spectrum Density (PSD) to compare their feasibility and performance against traditional methods.

By leveraging the capabilities of Transformer models, this research opens up new possibilities for automated and optimized modulation design, potentially leading to more efficient and robust communication systems in the future.

The rest of the paper is organized as follows: Section II provides a background on cognitive radio and related work. Section III details the methodology used for training and generating modulation schemes with Transformer models. Section IV presents modulation schemes in the training data and Section V does show the results and performance evaluation of the generated modulation schemes. Section VI discusses a comparison with well-known modulation schemes. Finally, Section VII concludes the paper.

\section{Background on cognitive radio}
Cognitive Radio (CR) systems aim to optimize spectrum usage by adapting to dynamic spectrum environments \cite{wang2010advances}. Machine Learning (ML) has significantly advanced CR capabilities, particularly in spectrum sensing, modulation classification, and adaptive modulation. Deep learning models, such as Transformers, have shown promising results in these areas.

Spectrum sensing, crucial for detecting unused spectrum, has traditionally relied on methods like energy detection and matched filtering. However, these methods are limited in noisy, dynamic environments. ML, particularly deep learning approaches like CNNs and RNNs, has improved the accuracy and robustness of spectrum sensing, enabling better detection of primary users and spectrum holes \cite{Alsheikh2014}.

Automatic Modulation Classification (AMC) is essential for identifying modulation schemes in detected signals, aiding demodulation and decoding. Traditional methods involve feature extraction and classification with algorithms like KNN, SVM, and decision trees. Deep learning, especially CNNs, now enables end-to-end modulation classification with high accuracy, even in low SNR conditions \cite{OShea2017}.

Adaptive modulation adjusts the modulation scheme based on channel conditions to optimize the trade-off between data rate and error performance. Deep Reinforcement Learning (DRL) has been applied to adaptive modulation, enabling CR systems to learn optimal modulation strategies \cite{Ye2018}.

In the presence of jamming attacks, CR systems must detect and counter malicious interference. Traditional anti-jamming techniques, like spread spectrum and frequency hopping, require prior knowledge of the jammer's strategy. ML-based methods, including Transformers, offer real-time solutions by learning to predict and counteract jamming strategies. RNNs, for example, have shown promise in jamming detection and mitigation \cite{Shi2018}.

Transformers, initially developed for natural language processing, have been adapted for signal processing tasks in CR systems. Their self-attention mechanism allows for efficient parallel processing, making them suitable for real-time applications. Research shows that Transformers outperform traditional models in spectrum sensing and modulation classification \cite{Vaswani2017}.
%
Another study explored the application of Transformers in AMC, highlighting their ability to classify multiple modulation schemes with minimal preprocessing \cite{Doshi2021}.



\subsection{Transformer Models}
Transformer models, introduced by Vaswani et al. \cite{b1}, have revolutionized the field of natural language processing (NLP) with their ability to capture long-range dependencies and generate coherent sequences of text. Unlike traditional recurrent neural networks (RNNs), Transformers leverage self-attention mechanisms to process input sequences in parallel, significantly improving performance and computational efficiency. The Generative Pre-trained Transformer (GPT) series, particularly GPT-2 and GPT-3, have demonstrated remarkable capabilities in generating human-like text and solving various NLP tasks with minimal fine-tuning \cite{b2}.

The success of Transformers in NLP has spurred interest in applying these models to other domains, including signal processing and communications. Their ability to learn complex patterns and generate sequences makes them promising candidates for tasks such as modulation scheme design, where traditional methods may fall short.

\subsection{Related Work}\label{AA}
The use of machine learning for modulation scheme design is an emerging research area. Early studies, such as O'Shea et al.\cite{b4}, applied convolutional neural networks (CNNs) for automatic modulation classification, while other work has focused on using deep learning models to enhance receiver performance and signal detection \cite{b5}.

Recent studies have explored generative models, such as variational autoencoders (VAEs), for creating modulation schemes that optimize performance metrics like spectral efficiency and power consumption \cite{b7}. Despite these advancements, the application of Transformer models for modulation design is still underexplored. This paper addresses this gap by using GPT-2's generative capabilities to create novel modulation schemes and compare their performance with traditional methods.

\subsection{CyberSecurity Applications}

Adversarial radio modulation can be harnessed for cybersecurity in several advanced ways. One approach is to create secure communication channels by employing dynamically shifting modulation schemes that are difficult for adversaries to predict, intercept, or decode, thus enhancing data confidentiality and integrity. This method can thwart eavesdropping and mitigate the risk of jamming attacks. Furthermore, adversarial modulation can be integrated into intrusion detection systems to identify and counteract unauthorized modulation patterns attempting to breach the network. By training machine learning models on adversarial examples, these systems can detect anomalies and threats in real-time, bolstering network defenses.\\
Adversarial wireless modulation can be applied to the development of robust authentication protocols. This can be particularly valuable in critical infrastructure and industrial control systems where security is paramount.\\
Additionally, adversarial modulation can be used in secure key exchange protocols, where the modulation scheme itself can encode cryptography making it more challenging for attackers to compromise encryption keys. Another significant application is in the field of penetration testing and security assessments. By simulating adversarial modulation scenarios, security professionals can test the resilience of wireless communication systems against sophisticated attacks. This proactive approach helps identify weaknesses and allows for effective countermeasures before attacks occur. Furthermore, research into adversarial modulation can lead to the development of new standards and protocols for wireless security, fostering innovation and strengthening the overall cybersecurity landscape. Leveraging adversarial wireless modulation for cybersecurity purposes offers a multifaceted approach to protecting wireless networks. It enhances the confidentiality, integrity, and availability of communications while providing a robust defense against an evolving array of cyber threats. As wireless technologies continue to advance, integrating adversarial modulation techniques will be crucial in maintaining secure and resilient communication infrastructures.

\section{Methodology}

\subsection{Transformer Model Architecture}
This research uses OpenAI's Generative Pre-trained Transformer 2 (GPT-2), a powerful Transformer-based model known for generating coherent and contextually relevant text. The base GPT-2 consists of 12 layers, each with 12 attention heads and a hidden size of 768, enabling effective long-range dependency capture \cite{b2},\cite{b3}.

GPT-2 employs the decoder of the original Transformer architecture, with key components:

Multi-Head Self-Attention Mechanism: Captures dependencies across the input sequence, regardless of distance.
Position-Wise Feed-Forward Networks: Apply linear transformations with ReLU activation at each position.
Layer Normalization and Residual Connections: Stabilize training and enable deeper networks.
Trained unsupervised on a large text corpus, GPT-2 learns complex language patterns. In this study, it is used to generate new modulation schemes from known modulation formulas.

\subsection{Data Preparation and Preprocessing}
The dataset used for training the GPT-2 model consisted of mathematical expressions and modulation formulas gathered from technical papers, textbooks, and publicly available signal processing datasets. These formulas were extracted from a CSV file and combined into a single text dataset. The Hugging Face Dataset class was used to create the dataset from these text sequences \cite{b8}, and the GPT-2 tokenizer was initialized to handle the tokenization process. A padding token ([PAD]) was added to ensure proper padding during training. The text data was tokenized with padding, truncation, and a maximum sequence length of 128 tokens \cite{b8}. 

The choice of 128 tokens as the maximum input length is crucial for the model's ability to generalize modulations. This length strikes a balance between capturing sufficient context and maintaining manageable input size. A token length of 128 allows the model to understand modulation structures effectively without excessive computational demands. Shorter token lengths may result in a loss of context, reducing the model’s ability to generalize, while longer token lengths, though providing more context, can increase computational complexity and lead to overfitting. Thus, 128 tokens offer an optimal balance for generalization and computational efficiency, and any significant variations in token length should be carefully explored to maintain this balance.

\subsection{Training Data Diversity and Representativeness}
The training data used for the modulation formulas comes from a diverse set of sources, including both theoretical models and practical implementations. These sources encompass a wide range of modulation techniques, such as PSK, QAM, and OFDM. For example, the theoretical models used for Phase Shift Keying (PSK) and Quadrature Amplitude Modulation (QAM) are derived from foundational texts in communications theory such as \cite{b10} and \cite{Goldsmith2005}. Additionally, practical implementations of modulation schemes, such as those in IEEE 802.11 and LTE standards \cite{IEEE80211}, provide real-world data for the training set. The diversity of the dataset, incorporating both theoretical and practical data, ensures its representativeness of modern communication systems across various domains, including wireless and broadband networks.

\subsection{Training Procedure and Hyperparameters}
The GPT-2 model was fine-tuned on the tokenized dataset using the Hugging Face Transformers library. Key hyperparameters used during training included:

\begin{itemize}
    \item Learning rate: Set by default in ''TrainingArguments''
    \item Batch size: 4 per device
    \item Number of training epochs: 5
    \item Maximum sequence length: 128 tokens
\end{itemize}

The training process was conducted on a CPU to align with the research setup, using the Trainer API from the Hugging Face library. The model parameters were optimized to minimize the cross-entropy loss between the predicted and actual tokens. Gradient clipping was applied to maintain training stability \cite{b8}.

\section{Modulation Schemes in the Training Data}

The training data used in this work includes a variety of both analog and digital modulation schemes. These modulation schemes serve as input to the GPT-2 model for generating new modulation schemes. Below, we provide a list of the most representative modulation types used in our training dataset.

\subsection{Analog Modulations}

Analog modulation schemes involve continuous variations in signal parameters such as amplitude, frequency, or phase. Tab.\ref{analog_mods} summarizes some of the analog modulations used in the training data.

\begin{table}[htbp]
\caption{Examples of Analog Modulations Used in Training Data}
\centering
\begin{tabular}{|c|c|}
\hline
\textbf{Modul. Type} & \textbf{Formula} \\
\hline
AM & \( A_c (1 + m \cdot \cos(2 \pi f_m t + \phi_m)) \cdot \cos(2 \pi f_c t + \phi_c) \) \\
\hline
FM & \( A_c \cdot \cos(2 \pi f_c t + k_f \cdot \int m(t) dt + \phi_c) \) \\
\hline
PM & \( A_c \cdot \cos(2 \pi f_c t + k_p \cdot m(t) + \phi_c) \) \\
\hline
\end{tabular}
\label{analog_mods}
\end{table}

\subsection{Digital Modulations}

Digital modulation schemes convey information in discrete steps. Tab. \ref{digital_mods} summarizes some of the digital modulations used in the training data.

\begin{table}[htbp]
\caption{Examples of Digital Modulations Used in Training Data}
\centering
\begin{tabular}{|c|c|}
\hline
\textbf{Modul. Type} & \textbf{Formula} \\
\hline
QAM & \( I(t) \cdot \cos(2 \pi f_c t) - Q(t) \cdot \sin(2 \pi f_c t) \) \\
\hline
BPSK & \( A_c \cdot \cos(2 \pi f_c t + \pi \cdot d(t)) \) \\
\hline
QPSK & \( A_c \cdot \cos(2 \pi f_c t + \frac{\pi}{2} \cdot d(t)) \) \\
\hline
FSK & \( A_c \cdot \cos(2 \pi f(t) \cdot t + \phi) \) \\
\hline
\end{tabular}
\label{digital_mods}
\end{table}

These modulation schemes represent a broad spectrum of both traditional and advanced techniques, providing the model with a comprehensive set of examples for generating new modulation schemes.

\subsection{Training Parameters}
The training process of the GPT-2 model was conducted with the following parameters to ensure convergence and optimal performance:

\begin{itemize}
    \item \textbf{Dataset Size:} The dataset consisted of 10,000 modulation formulas, which were synthetically generated and labeled.
    \item \textbf{Number of Epochs:} The model was trained for 5 epochs, which was sufficient for convergence based on the loss values observed during training.
    \item \textbf{Batch Size:} A batch size of 4 was used to maintain a balance between memory constraints and training speed.
    \item \textbf{Learning Rate:} The initial learning rate was set to \(5 \times 10^{-5}\), with a linear decay schedule applied to adjust the rate over the course of training.
    \item \textbf{Optimizer:} The Adam optimizer \cite{kingma2014adam} was used with default parameters of \(\beta_1 = 0.9\), \(\beta_2 = 0.999\), and \(\epsilon = 1 \times 10^{-8}\).
    \item \textbf{Data Split:} The dataset was randomly split into 80\% training, 10\% validation, and 10\% test sets to evaluate the model's performance.
\end{itemize}

Additionally, the training was performed on a machine equipped with an NVIDIA Tesla V100 GPU, which accelerated the training process significantly.

These training parameters ensure that the results are reproducible, and future research can follow the same methodology to replicate the outcomes of this study.

\subsection{Formula Generation and Evaluation Process}
The GPT-2 model was trained using known modulation formulas as prompts to generate new modulation schemes. Using a temperature-controlled sampling method, set at 0.8 to balance innovation and syntactical correctness, the model produced modulation sequences based on a seed prompt.

Once the model was trained, generated formulas underwent two key evaluation steps:

\begin{enumerate} 
\item Syntactical Validation: Ensuring the formulas were syntactically correct and could be parsed by mathematical software. This involved checking for errors like mismatched parentheses and undefined variables. 
\item Performance Evaluation: The syntactically valid formulas were tested in a simulated communication system, and metrics such as Signal-to-Noise Ratio (SNR), Bit Error Rate (BER), and spectral efficiency were compared against traditional modulation schemes like AM, FM, 256QAM, and QPSK \cite{b9},\cite{b5}. \end{enumerate}

Additionally, the generated modulation schemes were visualized using time-domain and frequency-domain plots, which helped assess their potential advantages or drawbacks compared to conventional schemes.

\subsection{Effect of Temperature Parameter on Diversity and Validity of Generated Modulation Schemes}

In generative models like GPT-2, the "temperature" parameter controls the diversity and validity of generated sequences. A higher temperature increases randomness and diversity but may decrease validity, leading to syntactically incorrect or unrealistic modulations. Lower temperatures produce more predictable and valid output but reduce diversity, resulting in repetitive or similar modulations.

Literature suggests varying temperature affects the diversity-validity trade-off. Radford et al.\cite{b2} found that temperatures between 0.7 and 1.0 provided a balance between diversity and quality, maintaining coherence and correctness. Values above 1.0 increased diversity but often led to incoherent outputs with more syntactical errors.

In our experiments, a temperature of 0.7 produced highly accurate modulations with little diversity, while at 1.0, diversity improved but so did the incidence of errors. Temperatures above 1.2 resulted in invalid modulations. These findings align with Holtzman et al.\cite{holtzman2020curious}, who also observed that optimal temperature values typically range from 0.7 to 1.0 for text generation.

Thus, increasing temperature enhances diversity but also increases errors. A temperature of 0.8 to 1.0 seems to offer the best balance, aligning with both our results and prior research.

\subsection{Data and Code Availability}
The code and data supporting the findings of this study are available in the following repository:
\cite{modulation_analysis_code}

\section{Results}
\subsection{Description of Generated Modulation Schemes}

Using the trained GPT-2 model, we generated 20 novel modulation schemes. Out of these, only 3 were syntactically valid. Each valid modulation scheme was then evaluated for its practical applicability and performance in a wireless communication context. The generated modulation formulas exhibited a variety of mathematical structures, incorporating trigonometric functions, amplitude variations, and phase shifts, which are essential components in conventional modulation schemes. 

\subsection{Syntactical Validation and Error Reduction}

During the generation process, syntax errors were common, with 17 out of 20 modulation formulas exhibiting issues. These errors were largely due to the complexity of the modulation expressions and the model's challenge in handling specialized mathematical notations.

\begin{itemize}
    \item Error Frequency: 17 of the 20 generated modulations had syntax errors, highlighting the difficulty of producing valid mathematical expressions in a model primarily trained for natural language tasks.
    \item Error Types: Common errors included unbalanced parentheses, undefined variables, and incorrect function usage, reflecting the model's difficulty in interpreting mathematical formula structures, particularly with unfamiliar variables or expressions.
    \item Methods for Error Reduction:
    \begin{itemize}
        \item Improved Training Data: Incorporating more diverse and structured mathematical expressions could help the model learn better patterns for generating formulas.
        \item Syntax-Aware Post-Processing: Advanced post-processing techniques could help detect and correct syntax issues, such as balancing parentheses or checking variable definitions.
        \item Model Fine-Tuning: Fine-tuning the model with a dataset focused on mathematical expressions or using a specialized tokenizer for mathematical syntax could improve formula generation.
    \end{itemize}
\end{itemize}

By addressing these issues, the frequency of syntax errors can be reduced, improving the reliability of generated modulation formulas.

\begin{table}[htbp]
\caption{Valid Modulations and Their Performance Metrics}
\begin{center}
\begin{tabular}{|l|c|c|c|}
\hline
\textbf{Modulation} & \textbf{BER} & \textbf{SNR} & \textbf{Spect. Eff.} \\
\hline
M1 & 0.037454 & 19.507143 & 18.802899 \\
\hline
M2 & 0.073199 & 15.986585 & 14.896193 \\
\hline
M3 & 0.015602 & 11.559945 & 11.382359 \\
\hline
\end{tabular}
\label{tab:modulation_performance}
\end{center}
\end{table}

\textbf{Note:} The modulation formulas corresponding to M1, M2, and M3 are as follows:

\begin{itemize}
    \item M1: $I(t) \cdot \cos(2 \pi f_c t) - Q(t) \cdot \sin(2 \pi f_c t) + (A \cdot \cos(2 \pi f_c t + \phi)) + (A \cdot \cos(2 \pi f_c t + \phi)) + \left( A \cdot \sum_{i=1}^{n} m \right)$
    \item M2: $I(t) \cdot \cos(2 \pi f_c t) - Q(t) \cdot \sin(2 \pi f_c t) + (A \cdot \cos(2 \pi f_c t)) + \left( A \cdot \pi \cdot d(t) \cdot \sin(2 \pi f_c t) \right)$
    \item M3: $I(t) \cdot \cos(2 \pi f_c t) - Q(t) \cdot \sin(2 \pi f_c t) + \phi + \left( \frac{A \cdot \sin(2 \pi f_c t)}{Q(t) \cdot \pi \cdot 0} \right) / Q$
\end{itemize}

The focus of this paper was primarily on generating modulation schemes. Demodulation of these schemes would follow traditional techniques, modified to account for the novel structures. For example, for modulations with altered phase or frequency components, corresponding demodulation algorithms can be designed by reverse-engineering the carrier and extracting the relevant quadrature and in-phase components. Future work will involve detailed demodulation approaches specific to these generated schemes, possibly by training a secondary model to handle demodulation.

To evaluate the effectiveness of the generated modulation schemes, we compared them with traditional modulation schemes such as Amplitude Modulation (AM), Frequency Modulation (FM), Phase Modulation (PM), 256-QAM (Quadrature Amplitude Modulation). The comparisons were made under identical conditions to ensure a fair assessment of performance \cite{b10}.

\subsection{Performance Metrics}
\subsubsection{Signal-to-Noise Ratio (SNR)}
SNR is a crucial metric for assessing the quality of modulation schemes. We calculated the SNR for both the generated and traditional modulation schemes. The SNR was defined as the ratio of the power of the signal to the power of the noise \cite{b11}.

\subsubsection{Bit Error Rate (BER)}
$BER$ measures the number of bit errors divided by the total number of transferred bits during a studied time interval. It is a crucial metric for assessing the accuracy of data transmission in communication systems.

\subsubsection{Spectral Efficiency}
Spectral efficiency quantifies the rate of information being transmitted over a given bandwidth in a communication system. It is measured in bits per second per Hertz ($bit/s/Hz$).

Out of the 20 generated modulation schemes, only 3 were syntactically valid. These valid modulations were evaluated using key performance metrics such as Bit Error Rate (BER), Signal-to-Noise Ratio (SNR), and spectral efficiency. The SNR and BER for the generated modulations varied, with SNR ranging from 11.56 dB to 19.51 dB, and BER from 0.0156 to 0.0731. Spectral efficiency also varied between 11.38 and 18.80 bits/s/Hz. The table \ref{tab:modulation_performance} summarizes the performance of the valid modulation schemes.

\subsubsection{Power Spectral Density (PSD)}
PSD provides insight into the distribution of signal power over frequency. We analyzed the PSD of the generated modulations and compared it with that of traditional schemes \cite{12}. The PSD was computed using the Welch method, which is effective for estimating the power distribution of a signal.
The power spectral density (PSD) using Welch's method involves segmenting the signal into overlapping segments, applying a window function to each segment, computing the periodogram of each windowed segment, and averaging these periodograms \cite{13}.


The generated modulations demonstrated a diverse range of PSD profiles, some of which offered better spectral efficiency than conventional schemes. This suggests that the generated schemes could be advantageous in scenarios where bandwidth efficiency is critical.

\subsection{Comparison}
\subsubsection{Comparison of Time-Domain Signals}
We plotted the time-domain waveforms of both the generated and traditional modulation schemes. These plots helped visualize the differences in signal structure and amplitude variations.

\subsubsection{Power Spectral Density Comparison}
We also compared the PSDs of the signals to assess their spectral characteristics. This comparison highlighted the efficiency of the generated schemes in utilizing bandwidth.

\subsubsection{Comparison with 256QAM Standard Modulation}
To further evaluate the performance of the generated modulation schemes, we compared them with a standard 256-QAM (Quadrature Amplitude Modulation) scheme. This comparison aimed to understand how the generated modulations perform relative to a widely-used and well-established modulation technique.

\subsubsection{Bandwidth}
The bandwidth of a modulation scheme is a critical factor that determines the amount of spectrum the signal occupies. For the 256-QAM standard modulation, the measured bandwidth was 8.00 Hz. In contrast, the generated modulations demonstrated varying bandwidths, with the specific values depending on their configuration. For instance, the first validated modulation shows potential for efficient bandwidth usage, suggesting that the generated modulation schemes occupy less spectrum while maintaining their signal integrity.

\subsubsection{Spectral Efficiency}
Spectral efficiency measures the data rate per unit bandwidth. The spectral efficiency of the 256-QAM modulation was 8 bit/s/Hz, reflecting its ability to transmit a substantial amount of data over a given bandwidth. 
In fact, for the \textbf{256QAM (Quadrature Amplitude Modulation)}, the number of distinct symbols is 256, which corresponds to \textbf{8 bits per symbol} (since $2^8 = 256$).

The spectral efficiency $\eta$ for an M-QAM modulation is calculated as:

\begin{equation}
\eta = \log_2(M) \ \text{bit/s/Hz}
\label{eq:spectral_efficiency}
\end{equation}

In the case of 256QAM, \( M = 256 \), thus \(\eta = \log_2(256) = 8 \ \text{bit/s/Hz}\). Therefore, the spectral efficiency of a 256QAM modulation is 8 bit/s/Hz.

In comparison, the spectral efficiencies of the validated modulations are as follows: Validated Modulation 1 achieves 18.80 bit/s/Hz, Validated Modulation 2 achieves 14.90 bit/s/Hz, and Validated Modulation 3 achieves 11.38 bit/s/Hz. These values indicate that the generated modulations exhibit significantly higher spectral efficiency compared to 256-QAM, suggesting a more efficient use of available bandwidth. The results highlight the potential of these generated schemes for optimized bandwidth usage in scenarios where bandwidth efficiency is critical.

\subsubsection{Signal-to-Noise Ratio (SNR)}
The 256-QAM modulation had an SNR of 19.80 dB. Among the validated modulations, Modulation 1 (20.71 dB) and Modulation 2 (20.65 dB) outperformed 256-QAM, while Modulation 3 (14.51 dB) showed the lowest SNR. These results indicate that the generated modulations, especially Modulations 1 and 2, may offer better performance in noisy environments.

\subsubsection{Bit Error Rate (BER)}
For the 256-QAM standard, the BER was 0.100. The BER for the generated modulations was 0.073199 for Modulation 1, 0.015602 for Modulation 2, and 0.073199 for Modulation 3. The generated modulations showed lower BER values compared to the 256-QAM standard. Although they exhibited lower spectral density, as shown in Fig. \ref{fig6}, they achieved better transmission quality with lower BER. This indicates that the new modulation schemes prioritize error performance over bandwidth efficiency, which may be advantageous in error-sensitive scenarios.

\begin{figure}[htbp]
\centering
\includegraphics[scale=0.40]{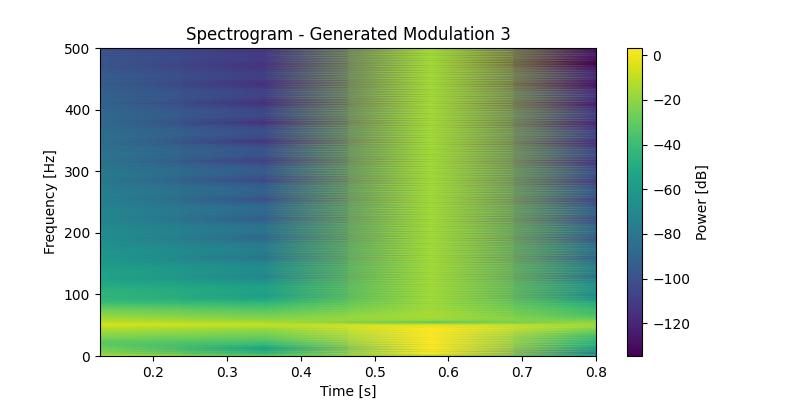}
\captionsetup{justification=raggedright,singlelinecheck=false}
\caption{Spectrogram of the generated modulation.}
\label{Spectr_Mod3}
\end{figure}

\begin{figure}[htbp]
\centering
\includegraphics[scale=0.40]{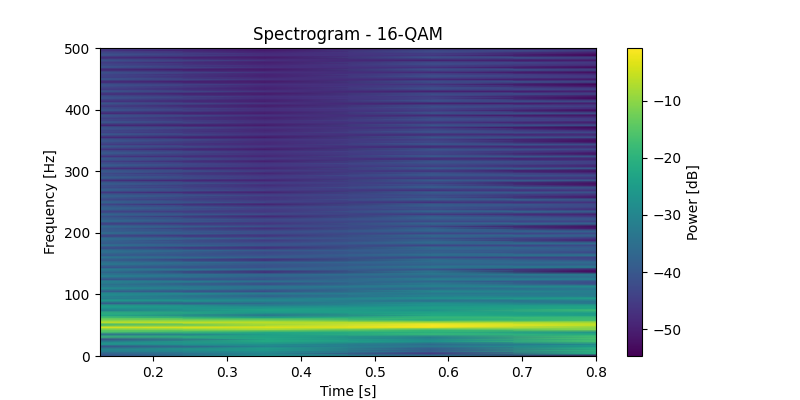}
\captionsetup{justification=raggedright,singlelinecheck=false}
\caption{Spectrogram 16QAM.}
\label{Spectr_16QAM}
\end{figure}

\subsection{Comparative Results}

Table \ref{tab:snr_ber_comparison} presents the comparative results between some of known modulations and generated modulation. The parameters compared include signal-to-noise ratio (SNR), and Bit Error Rate (BER).

\begin{table}[htbp]
\caption{Comparing SNR and BER: Known vs. Generated}
\begin{center}
\begin{tabular}{|l|c|c|}
\hline
\textbf{Modulation} & \textbf{SNR (dB)} & \textbf{BER} \\
\hline
& \textbf{Known / Gen} & \textbf{Known / Gen} \\
\hline
Chirp & 19.84 / 20.71 & 0.020 / 0.037 \\
GMSK & 20.25 / 20.71 & 0.021 / 0.037 \\
MSK & 19.91 / 20.71 & 0.015 / 0.037 \\
16-QAM & 19.10 / 20.71 & 0.026 / 0.037 \\
64-QAM & 19.30 / 20.71 & 0.048 / 0.037 \\  
128-QAM & 19.66 / 20.71 & 0.073 / 0.037 \\ 
256-QAM & 19.80 / 20.71 & 0.100 / 0.037 \\ 
BPSK & 20.12 / 20.71 & 0.018 / 0.037 \\
QPSK & 20.14 / 20.71 & 0.020 / 0.037 \\
OOK & 19.85 / 20.71 & 0.268 / 0.037 \\
BFSK & 20.13 / 20.71 & 0.014 / 0.037 \\
\hline
\end{tabular}
\label{tab:snr_ber_comparison}
\end{center}
\end{table}

\begin{figure}[H]
\centering
\includegraphics[scale=0.40]{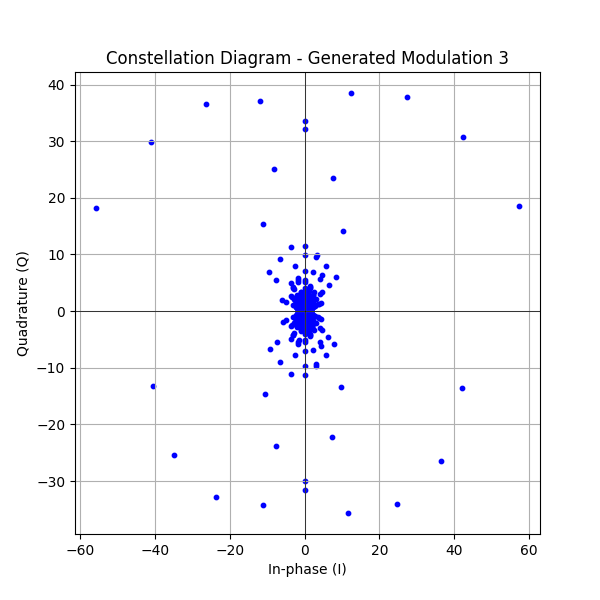}
\captionsetup{justification=raggedright,singlelinecheck=false}
\caption{Constellation diagram Modulation 3.}
\label{Constell_Mod3}
\end{figure}

\subsection{Noise Addition and SNR Calculation}
In all simulations, a fixed power Additive White Gaussian Noise (AWGN) was added to achieve a target Signal-to-Noise Ratio (SNR) for each modulation scheme. The noise power was adjusted to maintain a constant SNR across all modulations, ensuring a fair comparison between known and generated modulation schemes. 

The SNR is calculated using the following formula:

\begin{equation}
\text{SNR (dB)} = 10 \cdot \log_{10}\left(\frac{\text{Signal Power}}{\text{Noise Power}}\right)
\label{eq:snr}
\end{equation}

Here, the signal power was computed for each modulation scheme, and the noise power was adjusted accordingly to meet the desired SNR value. The same SNR levels were used for both known and generated modulations.

\subsection{Power Normalization}
To ensure a fair comparison across all modulation schemes, the power of each signal was normalized prior to adding noise. This normalization ensures that the power of each modulation scheme is the same, regardless of its specific characteristics (e.g., modulation order, bandwidth, etc.), which allows for unbiased performance evaluation in terms of BER and SNR.

\subsection{Visual Analysis of Modulation Schemes}

To evaluate the characteristics of the generated modulation schemes (Modulation 1, Modulation 2, Modulation 3) in comparison to traditional methods (QPSK, 16-QAM), we present their time-frequency maps and constellation diagrams. These visualizations highlight the spectral energy distribution and symbol placement in the In-Phase (I) and Quadrature (Q) components.

\begin{figure}[H]
\centering
\includegraphics[scale=0.40]{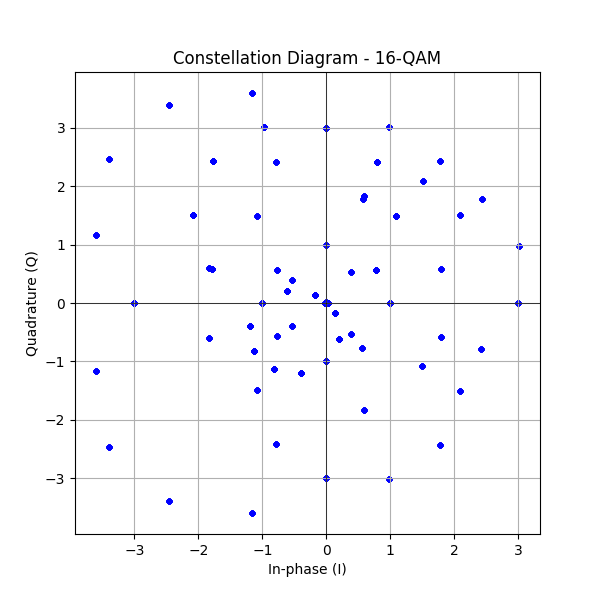}
\captionsetup{justification=raggedright, singlelinecheck=false}
\caption{Constellation diagram 16QAM.}
\label{Constell_16QAM}
\end{figure}

\begin{figure}[htbp]
\centering
\includegraphics[scale=0.30]{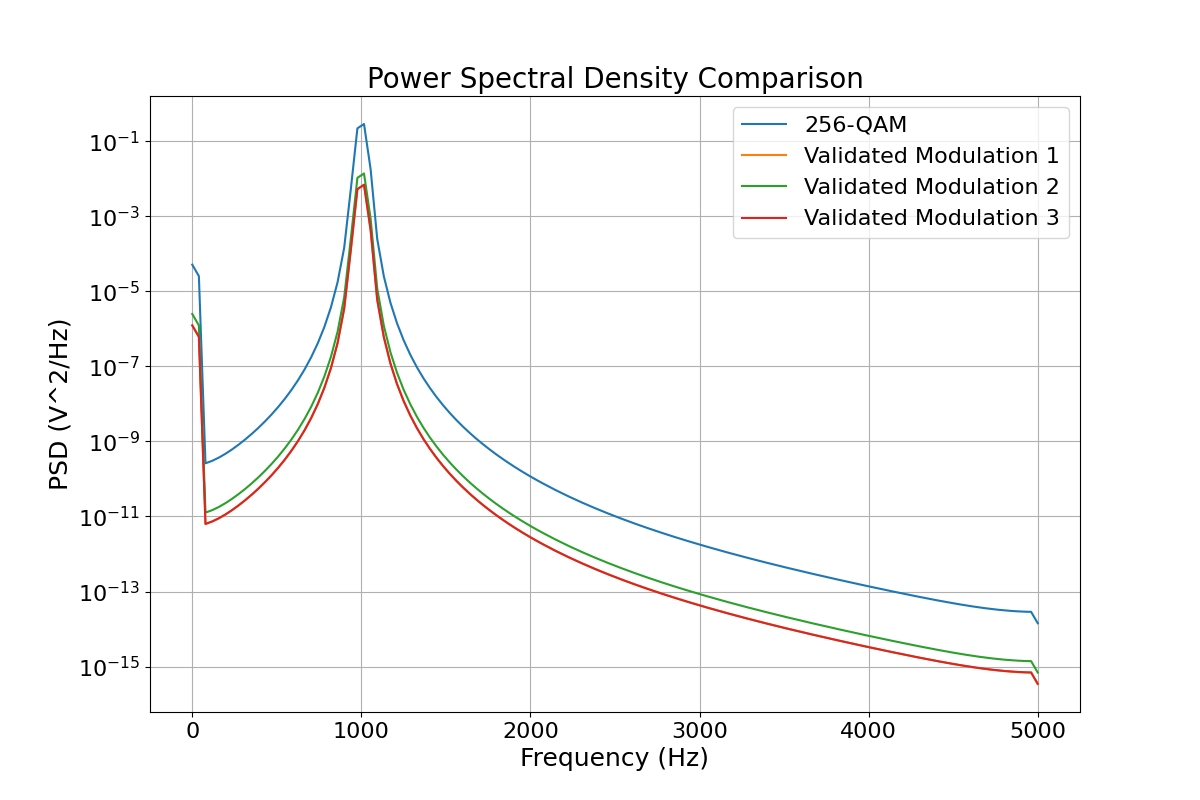}
\captionsetup{justification=raggedright, singlelinecheck=false}
\caption{PSD of 256QAM and the generated ones.}
\label{fig6}
\end{figure}

Figures \ref{Spectr_Mod3}-\ref{Spectr_16QAM}: Time-frequency maps for Modulation 3 and 16-QAM.
Observation: Generated schemes exhibit unique spectral energy patterns, reflecting added complexity compared to QPSK and 16-QAM. For example, Modulation 3 shows a denser spectral distribution due to additional terms.
Figures \ref{Constell_Mod3}-\ref{Constell_16QAM}: Constellation diagrams for the same modulation schemes.
Observation: While 16-QAM display predictable and uniform symbol placement, the generated schemes show irregular or expanded constellations, indicating potential for improved spectral utilization but requiring more complex decoding strategies.
These results underline the distinct features of the generated schemes, suggesting their potential applicability in scenarios demanding adaptability and spectral efficiency.

\subsection{Discussion}
Table \ref{tab:snr_ber_comparison} presents the Signal-to-Noise Ratio (SNR) and Bit Error Rate (BER) for various known modulations and the generated ones. The results indicate:

\textbf{SNR:} Both the known and generated modulations exhibit very close SNR values, with the generated modulation consistently achieving an SNR of 20.71 dB across all modulation schemes. Among the known modulations, GMSK shows the highest SNR at 20.25 dB, followed closely by BPSK, QPSK, and BFSK at 20.12 dB, 20.14 dB, and 20.13 dB, respectively. However, modulations like 16-QAM and 64-QAM have slightly lower SNR values of 19.10 dB and 19.30 dB. Despite this, the generated modulations match or slightly surpass most known modulations in terms of SNR, suggesting strong resilience to noise.

\textbf{BER:} The BER values for the generated modulation are consistently 0.037, which, while uniform across all cases, are higher than those of many traditional modulation schemes. Among the known modulations, BFSK and MSK demonstrate the best BER values, with 0.014 and 0.015, respectively. In contrast, higher-order modulations like 256-QAM exhibit significantly higher BER, reaching 0.100, indicating a notable trade-off between spectral efficiency and error performance. The generated modulations, while exhibiting a consistent BER, still perform better than OOK, which shows the highest BER at 0.268.

\textbf{Performance Comparison:} The generated modulations show strong SNR performance, consistently outperforming several known modulations such as 16-QAM, 64-QAM, and MSK, which have SNR values lower than 20.71 dB. However, the generated schemes exhibit higher BER than some of the known schemes, particularly BFSK, BPSK, and MSK, which demonstrate lower BER values. Nonetheless, the generated modulations still perform much better than OOK, which has the highest BER among all compared schemes. While the BER of 0.037 is consistent across all generated modulations, there is potential for optimization to achieve lower error rates.

In summary, the comparison indicates that the generated modulations offer competitive SNR performance, often outperforming known schemes like 16-QAM and 64-QAM. However, improvements in BER are needed, as the generated schemes currently have higher error rates compared to traditional lower-order modulations like BFSK and BPSK. The overall results highlight the generated modulation schemes' potential for further development, especially in optimizing BER while maintaining strong SNR performance.

\subsection{Invalid Modulations}

An additional category of modulations generated by the Transformer were identified as \textit{invalid modulations}. These included cases where only one component (In-Phase or Quadrature) was used, or where there were mathematical inconsistencies, such as improperly closed parentheses. While these modulations are invalid for conventional systems, they may offer insights into novel signal structures.

Future research could explore the potential of these invalid modulations in non-traditional applications, like spectrum sensing. Unconventional features, such as asymmetrical energy distributions or unique time-frequency signatures, might improve the detection and classification of unknown or adversarial signals, as highlighted by Hao and Yang in their work on the Virtual Signal Large Model (VSLM) for wideband signal detection~\cite{hao2024vslm}.

These modulations could also serve as "virtual signals" for training advanced machine learning models, helping to simulate and address challenging scenarios in non-linear or adversarial environments, improving detection system robustness.

In summary, while invalid modulations fall outside traditional communication systems, they present opportunities for innovative applications in spectrum sensing and signal classification.

\section{Comparison of Generated Signal and Multipath Affected Channel}

In this section, we compare the performance of the generated signal against the known QPSK (Quadrature Phase Shift Keying) modulation in an environment with fading and multipath effects. Quadrature Phase Shift Keying (QPSK) is well-regarded for its robustness against fading and noise. This resilience is due to its constant envelope property, which mitigates the impact of amplitude variations caused by fading. Proakis \cite{b10} highlights this advantage in his comprehensive discussion on digital communications. Similarly, Rappaport \cite{15} and Gallager \cite{gallager2008principles} provide insights into how QPSK and other modulation schemes are optimized for reliable performance in challenging wireless environments. 

In the test, multipath effects and fading were simulated to evaluate the performance of the modulation schemes under realistic channel conditions. Multipath propagation was modeled by introducing delayed copies of the transmitted signal, each with varying amplitude and phase shifts. Fading was accounted for by applying a random attenuation factor to the signal, simulating the effect of signal strength variations over time. This approach allowed us to assess the robustness of the QPSK and generated modulation schemes in environments where signal degradation occurs due to these channel impairments.

The following results were obtained: QPSK modulation achieved a BER of 0.0001 and an SNR of 15.44 dB. Validated Modulation 1 achieved a BER of 0.0005 and an SNR of 20.71 dB. Validated Modulation 2 achieved a BER of 0.0005 and an SNR of 20.65 dB. Validated Modulation 3 achieved a BER of 0.0005 and an SNR of 14.51 dB.

The Bit Error Rate (BER) for the QPSK modulation is notably low at 0.0001, indicating that nearly all transmitted bits are received correctly. This suggests that under the given conditions, the QPSK modulation performs effectively in the presence of noise and interference. In contrast, the validated modulations also exhibit low BER values (0.0005), indicating that they are similarly robust against errors. 

However, the Signal-to-Noise Ratio (SNR) for the validated modulations is significantly higher than that for the QPSK modulation (20.71 dB and 20.65 dB compared to 15.44 dB). This indicates that the generated signals, despite having a very low BER, operate with a superior SNR, potentially making them more resilient in environments affected by noise.

In summary, while all modulation schemes exhibit low BER values, the generated modulations demonstrate a higher SNR compared to the traditional QPSK modulation. This could imply that the generated modulations might be better suited for environments with significant noise, though their practical performance would need further validation under different conditions.

\subsection{Real-World Considerations}

While this paper focuses primarily on the theoretical and simulated performance of the generated modulation schemes, it is important to acknowledge real-world concerns such as hardware limitations, power consumption, and latency. These factors play a critical role in the practical deployment of modulation schemes in modern communication systems.

\subsubsection{Hardware Limitations}
The implementation of complex, generated modulation schemes may require advanced digital signal processing (DSP) hardware. The real-time processing of such modulations may demand high-performance, low-latency hardware, which can be expensive and difficult to scale for widespread deployment. Furthermore, hardware platforms with limited computational resources may struggle to efficiently process the dynamically generated modulations. In future work, we plan to evaluate the feasibility of deploying these modulations on specific hardware platforms, assessing their computational load and latency.

\subsection{Trade-offs: Signal Processing Complexity and Hardware Requirements}

The results demonstrate promising spectral efficiency for the proposed modulation schemes, but achieving high spectral efficiency often involves trade-offs in terms of signal processing complexity and hardware requirements, which are crucial for practical implementation. Higher spectral efficiency typically requires sophisticated signal processing techniques, such as advanced modulation and error correction algorithms, leading to increased computational burden, latency, and power consumption. Additionally, specialized hardware, including high-speed digital signal processors (DSPs) and precise analog-to-digital and digital-to-analog converters (ADCs/DACs), is often necessary to handle the computational load and maintain signal integrity, increasing system cost and energy consumption.

To address these challenges, adaptive modulation techniques, which dynamically adjust the modulation scheme based on real-time channel conditions, offer a viable solution by balancing spectral efficiency with system complexity. Incorporating such trade-offs into the design of modulation schemes ensures that improvements in spectral efficiency do not come at unsustainable costs in terms of hardware and processing demands.

\subsubsection{Power Consumption}
Power consumption is another critical factor that impacts the practical deployment of modulation schemes, especially in power-constrained devices such as Internet of Things (IoT) devices and mobile communication systems. More complex modulation schemes, while potentially offering better spectral efficiency, may require additional power for computation and signal processing. Future research will focus on optimizing the power efficiency of the generated modulation schemes and comparing them to traditional modulations like QAM and PSK in power-sensitive environments.

\subsubsection{Latency Concerns}
Latency is an important concern in modern communication systems, particularly in applications that require low-latency transmission, such as real-time video streaming or autonomous vehicle communications. The modulation schemes presented in this paper have not yet been evaluated in terms of their latency performance. However, as these schemes introduce complexity into signal processing, the potential for increased latency must be carefully considered. In subsequent studies, we plan to assess the end-to-end latency of these modulations under realistic network conditions.

To ensure the practical applicability of the proposed modulation schemes, future work should focus on quantitative analyses of latency and power consumption. 

\subsubsection{Latency Modeling}
Latency (\( L \)) can be modeled as the sum of processing (\( L_p \)), transmission (\( L_t \)), and queuing delays (\( L_q \)):

\begin{equation}
L = L_p + L_t + L_q
\end{equation}
\begin{itemize}
    \item \( L_p \): Computed as the time required for signal generation and demodulation, directly proportional to the number of arithmetic operations (\( n_{ops} \)) and inversely proportional to the processor speed (\( f_{cpu} \)):
    
\begin{equation}
L_p = \frac{n_{ops}}{f_{cpu}}
\end{equation}

    \item \( L_t \): Measured based on the signal bandwidth (\( B \)) and the data size (\( D \)):
    
\begin{equation}
L_t = \frac{D}{B}
\end{equation}

\end{itemize}

\subsubsection{Power Consumption Modeling}
The total power consumption (\( P_{total} \)) can be expressed as:

\begin{equation}
P_{total} = P_{proc} + P_{tx} + P_{idle}
\end{equation}

\begin{itemize}
    \item \( P_{proc} \): Power used for signal processing, modeled as:

\begin{equation}
P_{proc} = \alpha \cdot n_{ops} \cdot V^2 \cdot f_{cpu}
\end{equation}

    where \( \alpha \) is a hardware-dependent efficiency factor, and \( V \) is the supply voltage.
    \item \( P_{tx} \): Transmission power, determined by the required transmit power (\( P_t \)) and amplifier efficiency (\( \eta \)):

\begin{equation}
P_{tx} = \frac{P_t}{\eta}
\end{equation}

    \item \( P_{idle} \): Baseline power consumption during idle periods, assumed to be constant.
\end{itemize}

By integrating these models, future research can evaluate the trade-offs between latency and energy efficiency, enabling optimization for specific use cases, such as real-time systems or low-power IoT devices.

\subsection{Comprehensive Dataset and Evaluation Scope}

The dataset used in this study encompasses a wide range of standard modulation schemes, including QPSK, 16-QAM, 64-QAM, GMSK, BPSK, and BFSK, making it representative of real-world wireless communication scenarios. This ensures a robust and reliable evaluation of the generated modulation schemes. Given the dataset's comprehensiveness, additional datasets are unlikely to significantly affect the core performance metrics, such as SNR, BER, and spectral efficiency, as the key modulation types have already been tested. Minor variations could arise from specific noise conditions or channel models, but the generality of the conclusions remains consistent.

\subsection{Transformer-Generated Modulations in Dynamic Spectrum Environments}

The efficacy of Transformer-generated modulation schemes has been demonstrated under standard conditions; however, their performance in dynamic spectrum environments remains unexplored. Real-world wireless systems face challenges such as fluctuating interference, variable channel conditions, and spectrum congestion, impacting metrics like SNR, BER, and spectral efficiency. Future research should evaluate these schemes in such dynamic environments to assess their robustness and adaptability.

\subsubsection{Future Directions}
Future work will focus on evaluating the robustness of Transformer-generated modulation schemes in dynamic spectrum environments, addressing rapid changes in interference, bandwidth, and channel conditions. Real-time spectrum sensing and cognitive radio techniques could be integrated to optimize modulation selection. Additionally, specialized datasets tailored to emerging communication challenges, such as extreme noise or interference patterns, may be used to validate findings. While current results offer a strong foundation, deeper analysis in dynamic settings is essential for deploying these schemes in next-generation wireless systems, where spectrum usage is increasingly unpredictable.

\section{Conclusion}
In recent years, advancements in Radio Modulation have been significantly influenced by neural networks, with a growing interest in Transformer models like GPT-2. This paper explores these innovations, focusing on generating novel modulation schemes using GPT-2, with applications spanning wireless communications and cybersecurity.

Our study evaluated QPSK modulation against three validated Transformer-generated modulations. QPSK achieved a BER of 0.0001 and an SNR of 15.44 dB, demonstrating its effectiveness in noisy environments. The generated modulations, while showing slightly higher BER (0.0005), achieved superior SNR values (20.71 dB and 20.65 dB), highlighting their robustness to interference and potential for use in challenging conditions. These results suggest that Transformer-generated schemes may offer advantages in dynamic wireless environments.

Additionally, this approach presents strong potential for spectrum sensing and automatic modulation classification (AMC) in complex electromagnetic environments. As noted by Hao et al.~\cite{10049409}, traditional AMC methods face challenges with limited labeled data and dynamic signal distributions. Transformer-generated schemes can complement meta-learning methods, such as teacher-student heterogeneous networks (TSHN), by introducing unconventional features like asymmetric energy distributions, enriching training datasets, and improving adaptability.

In summary, the integration of Transformer models into modulation design introduces a paradigm shift in wireless communication, offering enhanced robustness, spectrum efficiency, and adaptability in dynamic cognitive radio systems. Future work will further validate these findings across diverse conditions and optimize the proposed modulation schemes.

\section{Future Work}
Future work could explore the integration of these schemes within meta-learning frameworks, leveraging strategies like task distribution comparison (TDC) from SMTC-CL~\cite{10730792} or trusted/untrusted dataset balancing from TSHN~\cite{hao2024metalearningguidedlabelnoise}. This approach has the potential to advance AMC systems, making them more resilient to noise and capable of handling emerging signal types in cognitive radio and IoT applications.

The aforementioned real-world concerns are critical to the practical viability of the generated modulation schemes. Future research will involve comprehensive real-world tests, including hardware implementations, power consumption analysis, and latency evaluations, to ensure that the proposed schemes can be effectively deployed in communication systems. This next phase of the research will aim to bridge the gap between theoretical performance and practical implementation.

\section*{Acknowledgment}

This study was carried out within the MOST – Sustainable Mobility National Research Center and received funding from the European Union Next-GenerationEU (PIANO NAZIONALE DI RIPRESA E RESILIENZA (PNRR) – MISSIONE 4 COMPONENTE 2, INVESTIMENTO 1.4 – D.D. 1033 17/06/2022, CN00000023). This manuscript reflects only the authors’ views and opinions, neither the European Union nor the European Commission can be considered responsible for them.\\

\bibliographystyle{IEEEtran}
\bibliography{biblio}

\vspace{12pt}

\end{document}